\documentclass[UTF8,a4paper]{article}
\usepackage{ctex}
\usepackage[sort&compress,numbers]{natbib}
\usepackage{caption} 
\usepackage{booktabs}
\usepackage{geometry}
\usepackage{titlesec}
\usepackage{url}
\usepackage[font={small,bf}]{caption}
\usepackage[yyyymmdd]{datetime}
\usepackage{graphicx}
\usepackage{amsmath}
\usepackage{makecell}
\usepackage{multirow}
\usepackage{array}
\usepackage{color}
\usepackage{threeparttable}

\titleformat*{\section}{\songti\zihao{4}\bfseries}
\titleformat*{\subsection}{\heiti\zihao{5}\bfseries}
\titleformat*{\subsubsection}{\kaishu\zihao{5}\bfseries}

\newenvironment{csmtAbstract}{\noindent \kaishu \small {\bfseries Abstract:}}{}
\newenvironment{keywords}{\small \noindent{\bfseries Key Words:}}{}

\newcommand{\weizhu}[1] {\noindent{\bfseries Recieved Date:~\today} \newline {\bfseries \indent ~~Author:} #1}

\newcommand*{\affaddr}[1]{{\sffamily \small (\textit{#1})}} 
\newcommand*{\affmark}[1][*]{\textsuperscript{#1}}

\makeatletter
\renewcommand\@maketitle{%
	\hfill
	\begin{center}
		\vskip 2em
		\let\footnote\thanks 
		{\centering \bfseries \zihao{3} \@title \thanks{\weizhu{Fan Bu (2001--), M.Eng. (Research) Student, m.july@qq.com; Rongfeng Li (1984--), Associate Professor.}} \par } 
		\vskip 1em
		{\centering \zihao{5} \textbf{\@author} \par}
	\end{center}
	\vskip 1em \par
}
\makeatother

\title{The Renaissance of Expert Systems:\\Optical Recognition of Printed\\Chinese \textit{Jianpu} Musical Scores with Lyrics}

\author{%
	{Fan Bu\affmark[1], Rongfeng Li\affmark[1], Zijin Li\affmark[2], Ya Li\affmark[3], Linfeng Fan\affmark[2], Pei Huang\affmark[1]}\\
	\affaddr{\affmark[1]Beijing University of Posts and Telecommunications,\\\affmark[2]Central Conservatory of Music, \affmark[3]Shanghai Normal University}\\
}

\begin{document}
	
    \maketitle

    \vspace{-5pt}
    
    \begin{csmtAbstract}
        Large-scale optical music recognition (OMR) research has focused mainly on Western staff notation, leaving Chinese \textit{Jianpu} (numbered notation) and its rich lyric resources underexplored. We present a modular expert-system pipeline that converts printed \textit{Jianpu} scores with lyrics into machine-readable MusicXML and MIDI, without requiring massive annotated training data. Our approach adopts a top-down expert-system design, leveraging traditional computer-vision techniques (e.g., phrase correlation, skeleton analysis) to capitalize on prior knowledge, while integrating unsupervised deep-learning modules for image feature embeddings. This hybrid strategy strikes a balance between interpretability and accuracy. Evaluated on \textit{The Anthology of Chinese Folk Songs}, our system massively digitizes (i) a melody-only collection of more than 5,000 songs (> 300,000 notes) and (ii) a curated subset with lyrics comprising over 1,400 songs (> 100,000 notes). The system achieves high-precision recognition on both melody (note-wise F1 = 0.951) and aligned lyrics (character-wise F1 = 0.931).
    \end{csmtAbstract}
    

    \vspace{-5pt}

    \section{Introduction}

    Music digitization is a key task in computational musicology, driving progress in analysis, retrieval, and generative tasks. Large-scale datasets like The Harmonix Set \cite{the_harmonix_set} (912 annotated pop songs) and IrishMAN \cite{irishman} (over 216,000 Irish melodies) have greatly advanced these fields. However, most resources focus on Western pop and classical music, leaving Chinese traditional and folk music—rich in cultural, linguistic, and performative diversity—underrepresented. Existing collections, such as Essen's Chinese folksong subset \footnote{Chinese subset, the Essen folksong collection: \url{https://kern.humdrum.org/cgi-bin/browse?l=essen/asia/china/}} (2,179 melodies), often lack key elements like lyrics and are insufficient in volume for modern, data-intensive models.

\begin{figure}[t!]
        \centering
        \includegraphics[width=0.55\columnwidth]{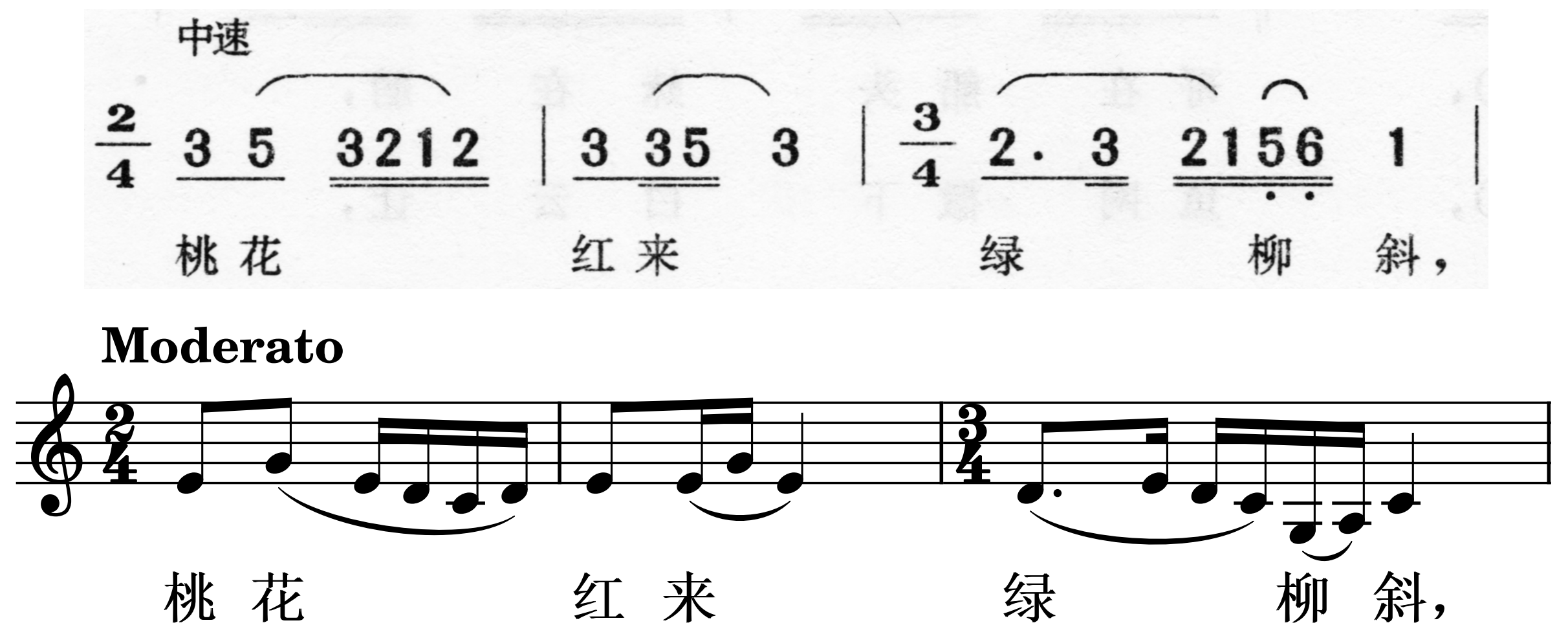}
        \vspace{-8pt}
        \caption{A phrase of \textit{Jianpu} and the equivalent staff notation.}
        \label{fig:fig01_jianpu}
    \end{figure}

    A major challenge lies in the notation format. While Western staff notation dominates in Optical Music Recognition (OMR) systems (e.g., DeepScores \cite{deepscores}, DoReMi \cite{doremi} for printed scores and MUSCIMA++ \cite{muscima_plus_plus} for handwritten), Chinese music is typically transcribed in \textit{Jianpu} (numbered notation). \textit{Jianpu} encodes pitch with digits and octave-shifting dots, and indicates duration with underlines and dashes (see Fig.~\ref{fig:fig01_jianpu}), requiring specialized OMR pipelines since staff notation priors are of limited use.

    \vspace{-12pt}

    Despite the need for large, annotated \textit{Jianpu}+lyric OMR datasets, their scarcity limits the application of modern machine learning techniques. Some pioneering attempts have created \textit{Jianpu} OMR datasets using rendered or randomly generated scores \cite{wang_qi_jianpu_omr} \cite{bu_fan_jianpu_omr} to train CNNs or object detectors like YOLO \cite{yolo}. While promising, these efforts suffer from accuracy issues, as a single transcription error in note duration can invalidate an entire phrase.
    
\begin{figure}[h!]
        \centering
        \includegraphics[width=1.0\columnwidth]{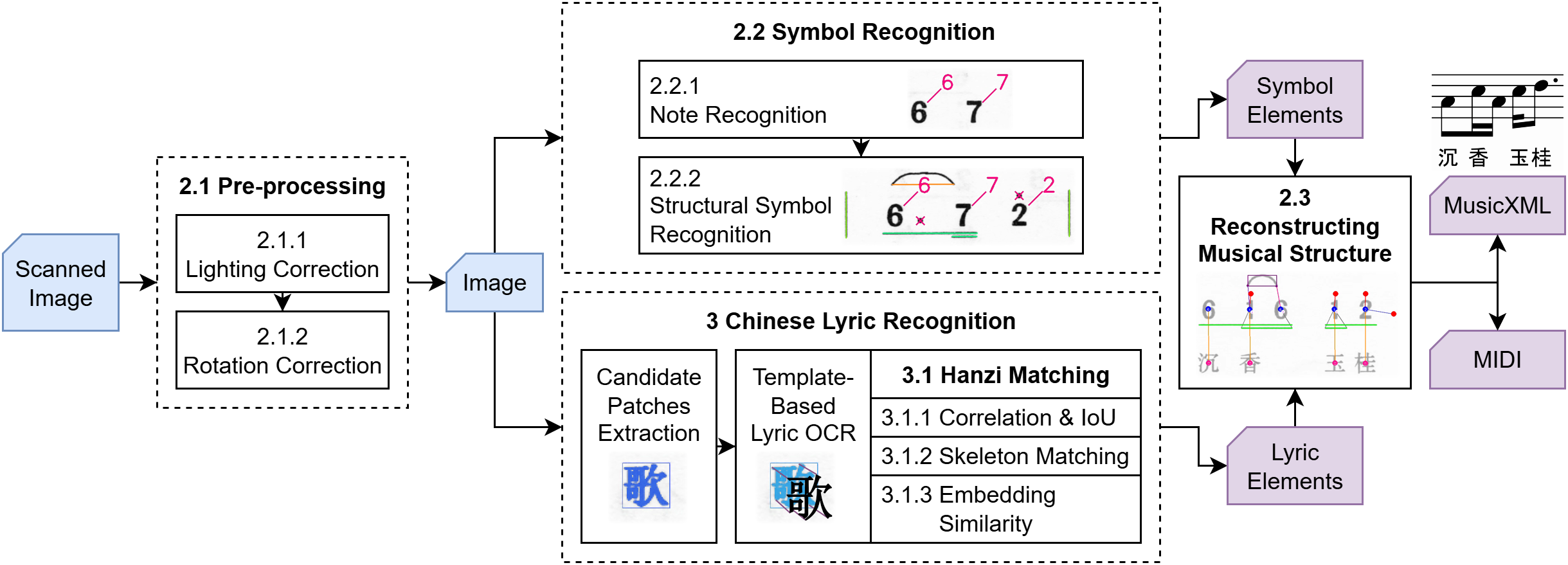}
        \vspace{-15pt}
        \caption{A diagram of our \textit{Jianpu} OMR pipeline.}
        \label{fig:fig00_overall}
    \end{figure}

    To address these gaps, we propose a modular, expert-system-based recognition pipeline (see Fig.~\ref{fig:fig00_overall}) for printed \textit{Jianpu} scores with lyrics. Our largely unsupervised, top-down approach achieves high accuracy without extensive data. Section 2 discusses the symbol recognition process, which uses phrase correlation and skeleton-based geometric analysis to handle \textit{Jianpu} elements. Section 3 introduces a Chinese lyric OCR module combining template matching and unsupervised metric-learning. Section 4 demonstrates the pipeline’s effectiveness in the large-scale digitization of \textit{The Anthology of Chinese Folk Songs}, generating over 5,000 songs with high recognition accuracy.

    \vspace{-15pt}

    \section{\textit{Jianpu} Melody Recognition}

    \subsection{\textit{Jianpu} Image Preprocessing}

    Aimed to enhance the quality of scanned \textit{Jianpu} images, we employed adaptive lighting and rotation correction. These techniques improve both the visual clarity and structural integrity of the scanned images, preparing them for subsequent analysis.

    \subsubsection{Lighting Correction: Dual Gamma Transform}

    Lighting correction is essential for standardizing the non-uniform illumination found in scanned images. Our method first estimate background and foreground intensities ($v_{\rm BG}$ and $v_{\rm FG}$), then apply a dual-gamma transform to normalize them.

    \textbf{Background.} \textit{Jianpu} layout typically has a background proportion exceeding 75\%. With PDF $h(v)$ and CDF $H(v)$ ($v\!\in\![0,1]$) of the image grayscale values, the background cutoff $v_{\rm BG\uparrow}$ is the quantile $\alpha\!=\!0.75$ and the background intensity $v_{\rm BG}$ is yielded as in Eq.~\ref{eq:1}:

    \vspace{-6pt}
    \begin{equation}
        v_{\rm BG} = \frac {\int_0^{v_{\rm BG \uparrow}} v \cdot h(v) {\rm d} v}{\int_0^{v_{\rm BG \uparrow}} h(v) {\rm d} v}, \quad v_{\rm BG \uparrow} = H^{-1}(\alpha)
        \label{eq:1}
    \end{equation}
    
    \textbf{Foreground.} The foreground proportion varies depending on the printed content. Thus the foreground cutoff $v_{\rm FG \downarrow}$ uses Otsu's thresholding method, as in Eq.~\ref{eq:2}: 

    \vspace{-6pt}
    \begin{equation}
        \quad v_{\rm FG} = \frac {\int_{v_{\rm FG \downarrow}}^1 v \cdot h(v) {\rm d} v}{\int_{v_{\rm FG \downarrow}}^1 h(v) {\rm d} v}, \quad v_{\rm FG \downarrow} = {\rm Otsu} (f)
        \label{eq:2}
    \end{equation}
    
    \textbf{Dual-gamma transform.} Finally, we apply dual gamma transform (Eq.~\ref{eq:3}) to the image, normalizing estimated $v_{\rm BG}$ and $v_{\rm FG}$ to target intensities $v_{\rm BGT} = 0.01$ and $v_{\rm FGT} = 0.9$:

    \vspace{-5pt}
    \begin{equation}
        f' = 1 - (1 - f^{\gamma_1})^{\gamma_2}, \quad \gamma_1 = \frac {\ln v_{\rm BGT} }{\ln v_{\rm BG}}, \quad \gamma_2 = \frac {\ln (1 - v_{\rm FGT})}{\ln (1 - v_{\rm FG})}
        \label{eq:3}
    \end{equation}
    
\begin{figure}[h!]
        \centering
        \includegraphics[width=\columnwidth]{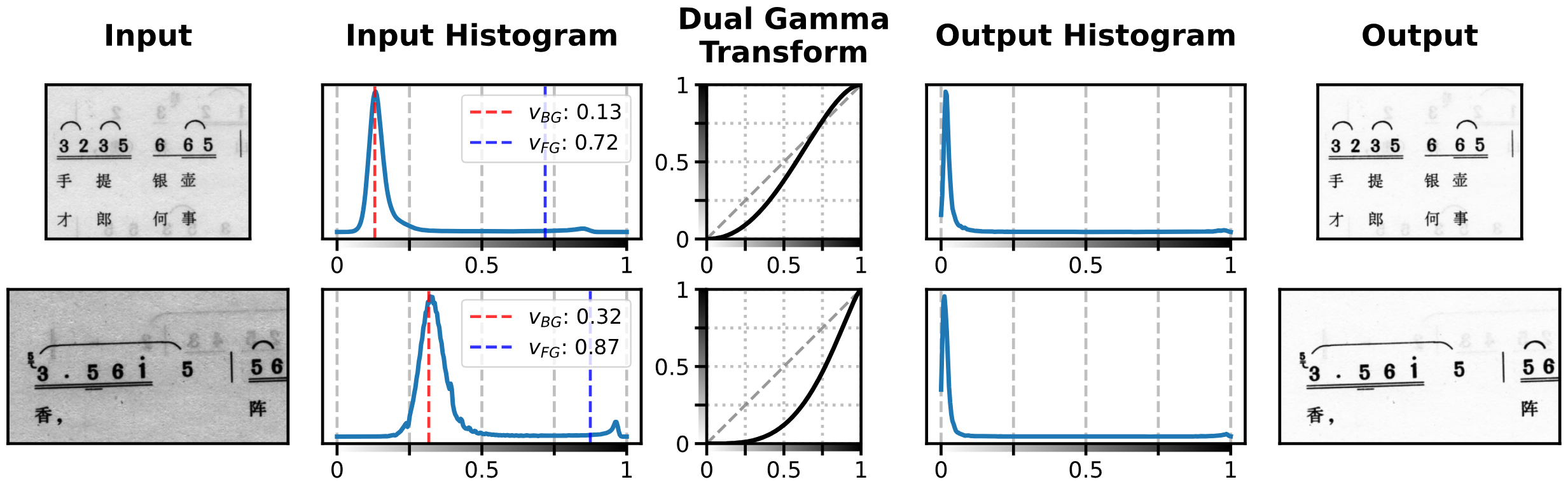}
        \vspace{-16pt}
        \caption{Examples of the adaptive lighting correction process using dual gamma transform.}
        \label{fig:fig02_dual_gamma}
    \end{figure}

    As illustrated in Fig.~\ref{fig:fig02_dual_gamma}, the application of this algorithm yields images with a consistent lighting pattern, irrespective of their initial illumination conditions and content.

    \vspace{-10pt}
    
    \subsubsection{Rotation Correction: Entropy-minimization}

    Image skew is corrected by finding the angle $\theta$ that minimizes the normalized Shannon entropy of the horizontal projection profile. For an image $f[i,j;\theta]$, with height $N$, width $M$) and rotated by $\theta$, the entropy to optimize is defined in Eq.~\ref{eq:4}:

    \vspace{-16pt}
    \begin{equation}
        H(\theta) = -\frac {1}{\log_2 N} \sum_{i=0}^{N-1} p[i; \theta] \log_2 p[i; \theta], \quad p[i; \theta] = \frac {b[i; \theta]}{\sum_{i'=0}^{N-1}b[i'; \theta]}, \quad b[i; \theta] = \sum_{j=0}^{M-1} f[i, j; \theta]
        \label{eq:4}
    \end{equation}
    
\begin{figure}[h!]
        \centering
        \includegraphics[width=0.7\columnwidth]{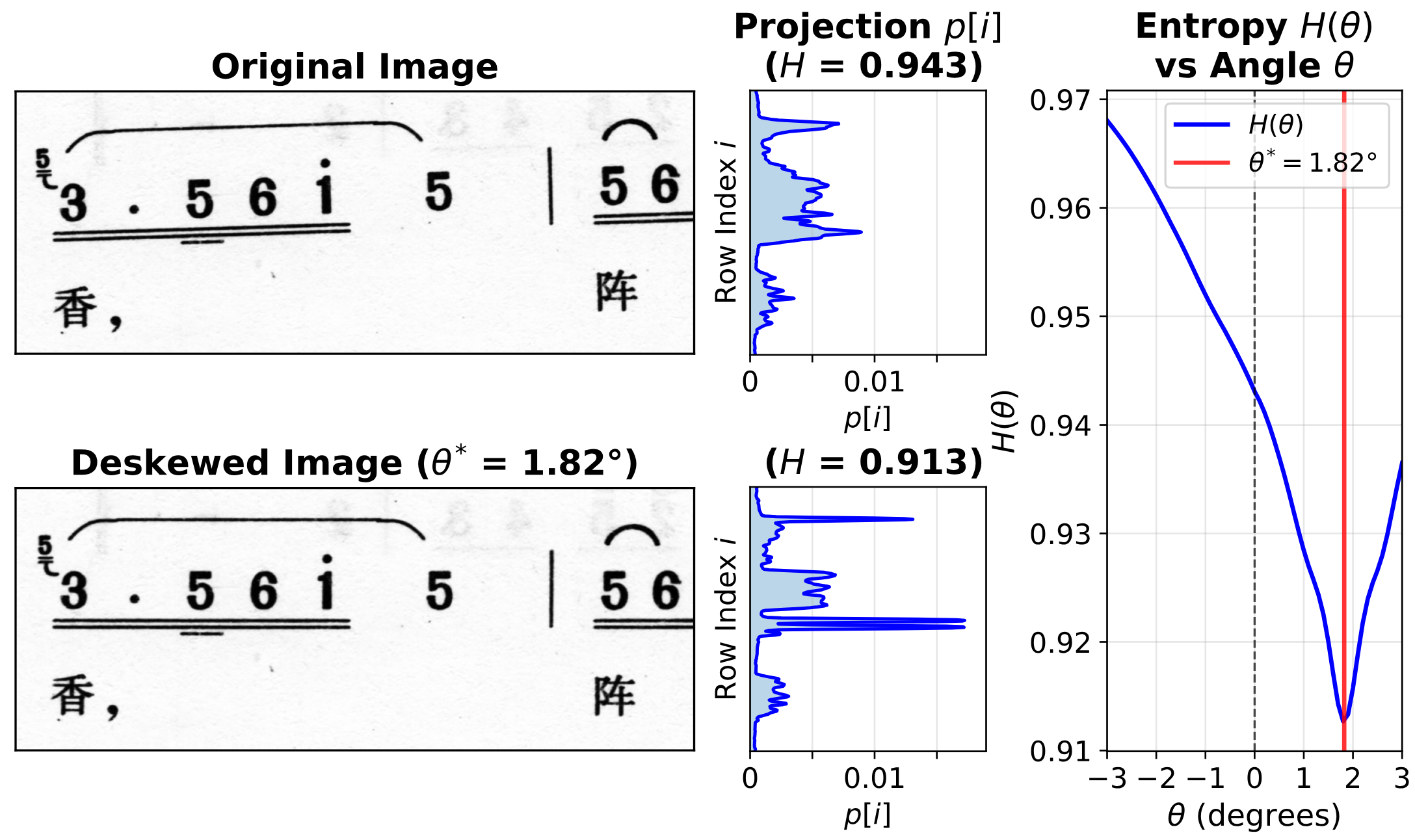}
        \vspace{-8pt}
        \caption{Demonstration of entropy-minimization rotation correction.}
        \label{fig:fig03_deskew}
    \end{figure}

    The entropy function $H(\theta)$ is typically unimodal for a \textit{Jianpu} section or an entire page, and a smaller value indicates clearer row separation and hence better alignment. (see Fig.~\ref{fig:fig03_deskew}). Consequently, the optimal angle can be determined using the golden-section search algorithm. To expedite convergence, a multi-scale Gaussian pyramid reduction is employed, where initial steps are solved on a low-resolution version of the image.

    \vspace{-8pt}

    \subsection{\textit{Jianpu} Symbol Recognition}

    Symbol recognition is a critical step in the \textit{Jianpu} OMR system, converting visual music representations into machine-readable formats.

    \vspace{-4pt}
    
    \subsubsection{Note Recognition: Template Correlation Response Analysis}

    In the same \textit{Jianpu} publication, notes (digits 1–7) and rests (digit 0) are printed in uniform font style and size. We employ handcrafted correlation templates to detect digits.

\begin{figure}[h]
        \centering
        \includegraphics[width=0.8\columnwidth]{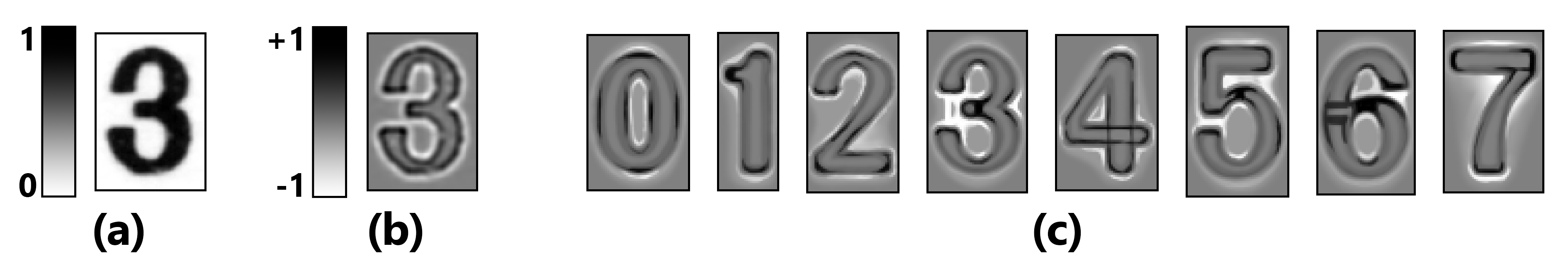}
        \vspace{-6pt}
        \caption{(a) An example digit image; (b) LoG-filtered; (c) manually enhanced templates.}
        \label{fig:fig04_digit_templates}
    \end{figure}

    For every digit, we extract a representative glyph, apply a Laplacian-of-Gaussian (LoG) filter, and manually add accents to enhance the distinction between background and foreground, especially where similar digits intersect. All digit templates are presented in Fig.~\ref{fig:fig04_digit_templates}.

    \vspace{-8pt}

    \vspace{-6pt}

    Digit recognition is performed by calculating the correlation between the digit template and the score image, followed by thresholding high-response regions. Although seemingly straightforward, the method proves highly effective: as detailed in Section 4.3, it attains 100\% accuracy across 233 evaluated note samples.

    \vspace{-4pt}

    \subsubsection{Structural Symbol Recognition: Skeleton Analysis}

    Structural symbols, such as dots, underlines, dashes, barlines, ties, and slurs, are recognized through morphological and skeleton analysis. To illustrate, we only describe the algorithm for detecting ties and slurs.

\begin{figure}[h!]
        \centering
        \includegraphics[width=0.35\columnwidth]{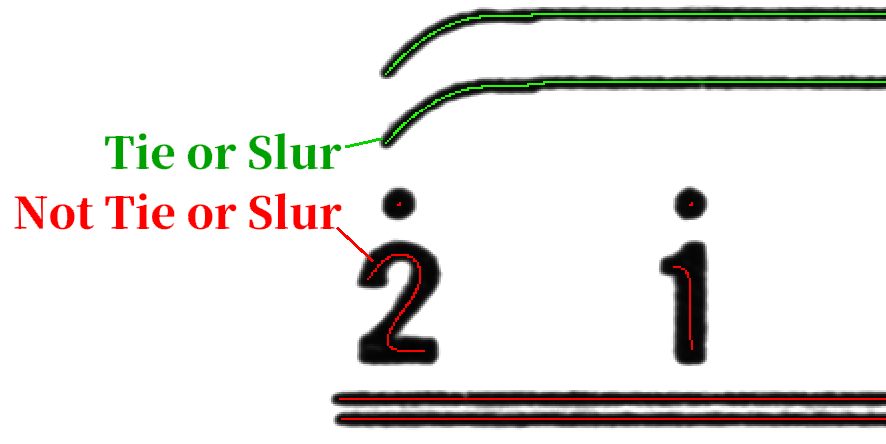}
        \vspace{-6pt}
        \caption{Example of tie and slur recognition.}
        \label{fig:fig05_tieslur_detection}
    \end{figure}

    First, morphological operations are used to enhance the image, which can eliminate unwanted artifacts and fix broken strokes. Connected components are extracted from the skeleton image generated using Zhang-Suen algorithm \cite{zha84}. For each component, a graph model is created, with pixels as vertices and M-adjacency relations as edges. The longest chain within each component, identified via breadth-first search (BFS), represents the component. The chain is smoothed, analyzed, and classified as a musical tie or slur if it satisfies specific geometric constraints, such as length and directionality (see Fig.~\ref{fig:fig05_tieslur_detection}).

    \vspace{-6pt}

    \vspace{-16pt}

    \subsection{Reconstructing Musical Structure: Anisotropic Relationship Analysis}

\begin{figure}[h!]
        \centering
        \includegraphics[width=0.72\columnwidth]{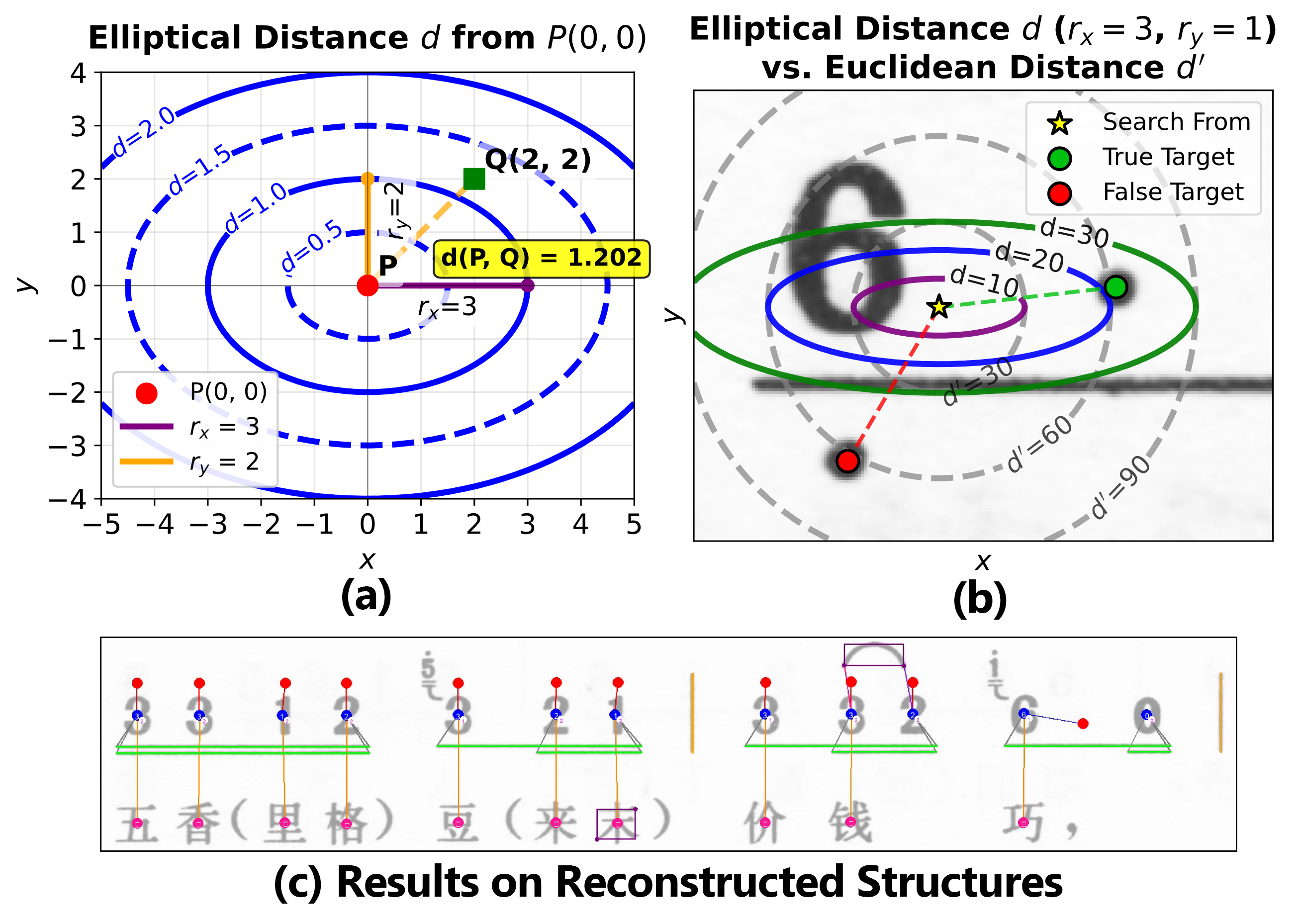}
        \vspace{-5pt}
        \caption{Anisotropic elliptical distance for relationship extraction between musical symbols.}
        \label{fig:fig06_relation}
    \end{figure}

    Relationships between \textit{Jianpu} elements are spatially anisotropic. Information related to pitch and performance techniques typically appears along the vertical axis, while timing-related information is predominantly aligned horizontally. Isotropic Euclidean nearest-neighbor search methods often fail to differentiate between elements that are horizontally or vertically related. To solve this, we introduce an anisotropic elliptical distance function $d(\cdot)$, as defined in Eq.~\ref{eq:6} and illustrated in Fig.~\ref{fig:fig06_relation}(a):

    \vspace{-2pt}

    \begin{equation}
        d(\boldsymbol p, \boldsymbol q; r_x, r_y) = \sqrt {\left ( \frac {p_x - q_x}{r_x} \right )^2 + \left ( \frac {p_y - q_y}{r_y} \right )^2 }
        \label{eq:6}
    \end{equation}

    \vspace{-8pt}

    Fig.~\ref{fig:fig06_relation}(b) illustrates an example where the Euclidean distance incorrectly identifies the octave-shifting dot below as an augmentation dot, whereas our elliptical distance function performs correctly.
    
    This distance function is incorporated into a balanced KD-tree index. Unlike standard KD-trees\cite{kd_tree}, which assume Euclidean space, our method supports dynamic anisotropic parameters $(r_x, r_y)$ during query execution without needing to rebuild the index. This is accomplished by scaling original pruning operations with $s_x = 1/r_x$ and $s_y = 1/r_y$ on the fly. As a result, queries can be processed in $O(\log n)$ time for well-distributed data, similar to traditional KD-trees.
    
    This spatial index enables efficient and robust extraction of complex semantic relationships, such as octave shifts, duration adjustments, and lyric alignment (see Fig.~\ref{fig:fig06_relation} (c)). The extracted musical structure can be easily converted into MusicXML and MIDI format.

    \vspace{-4pt}

    \section{Localization and Recognition of Chinese Lyrics}

    Recognizing Chinese lyrics in printed \textit{Jianpu} scores presents unique challenges. Unlike typical OCR tasks, where characters are grouped into words or regions, Chinese characters (\textit{Hanzi}) in \textit{Jianpu} are often scattered across the page and spatially aligned with musical notes. This requires character-level identification. Modern OCR algorithms, which generally assume text is organized into words or larger regions (a structure common in English), face difficulties in this context. In this section, we propose a robust solution for character-wise Chinese lyric localization and recognition.

    \vspace{-4pt}

    \subsection{Unsupervised \textit{Hanzi} Matching}
    
    Given the fixed font style and size within a publication, we employ an unsupervised template matching approach, combining multiple similarity metrics to measure the similarity between an extracted candidate patch $f_1$ and a template patch $f_2$.

    \vspace{-4pt}
    
    \subsubsection{Scale-Optimized Phase Correlation: Translation and Scaling Alignment}

    We apply phase correlation to estimate translations between character patches, using the Fourier shift theorem to leverage the FFT algorithm, as shown in Eq.~\ref{eq:7}:

    \vspace{-6pt}

    \begin{equation}
            R(u,v) = \frac{F_1(u,v) \cdot F_2^*(u,v)}{|F_1(u,v) \cdot F_2^*(u,v)|}, \quad C(x,y) = \mathcal{F}^{-1}\{R(u,v)\}
        \label{eq:7}
    \end{equation}

    The peak of $C(x,y)$ provides the translation offset $(t_x,t_y)$. Since even small scaling differences can cause significant displacement in \textit{Hanzi} strokes, we extend the correlation method with a golden section search over scale factors. After translation and scaling alignment, we calculate the normalized correlation response and a min-max grayscale Intersection over Union (IoU) metric to assess similarity.

    \vspace{-4pt}
    
    \subsubsection{Skeleton Matching: Structural Robustness}

    Character patches are reduced to skeleton point sets, with similarity quantified as a minimum-cost bipartite matching between the two point sets $\{ \boldsymbol a_i \}_{i=1}^n$ and $\{ \boldsymbol b_j \}_{j=1}^m$. The cost is defined as squared L2 distance between matched points, with a penalty applied to unmatched points, which is equivalent to a distance cost when $\| \boldsymbol a_i - \boldsymbol b_j \|_2 = \lambda$, where $\lambda$ is chosen based on font size (typically $\lambda=12$). The cost matrix is defined as in Eq.~\ref{eq:8}:

    \vspace{-6pt}

    \begin{equation}
        \boldsymbol C = \begin{bmatrix} \boldsymbol D & \boldsymbol \Lambda_{n \times n} \\ \boldsymbol \Lambda_{m \times m} & \boldsymbol D^{\rm T} \end{bmatrix}, \quad d_{i,j} = \frac 1 2 \| \boldsymbol a_i - \boldsymbol b_j \|_2^2, \quad \boldsymbol \Lambda = \frac 1 2 \lambda^2 \cdot {\rm diag}\, \boldsymbol 1
        \label{eq:8}
    \end{equation}
    
    The optimal matching is obtained using the Hungarian algorithm, and the normalized similarity metric is derived from the corresponding cost $J^*$ as in Eq.~\ref{eq:9}:

    \vspace{-6pt}

    \begin{equation}
        s_{\rm Skeleton} = \exp \left ( - \frac {J^*}{\lambda^2 \cdot (\frac {m+n}{2})} \right )
        \label{eq:9}
    \end{equation}
    
    \subsubsection{Unsupervised Embedding Similarity: Semantic Robustness}

\begin{figure}[h!]
        \centering
        \includegraphics[width=0.6\columnwidth]{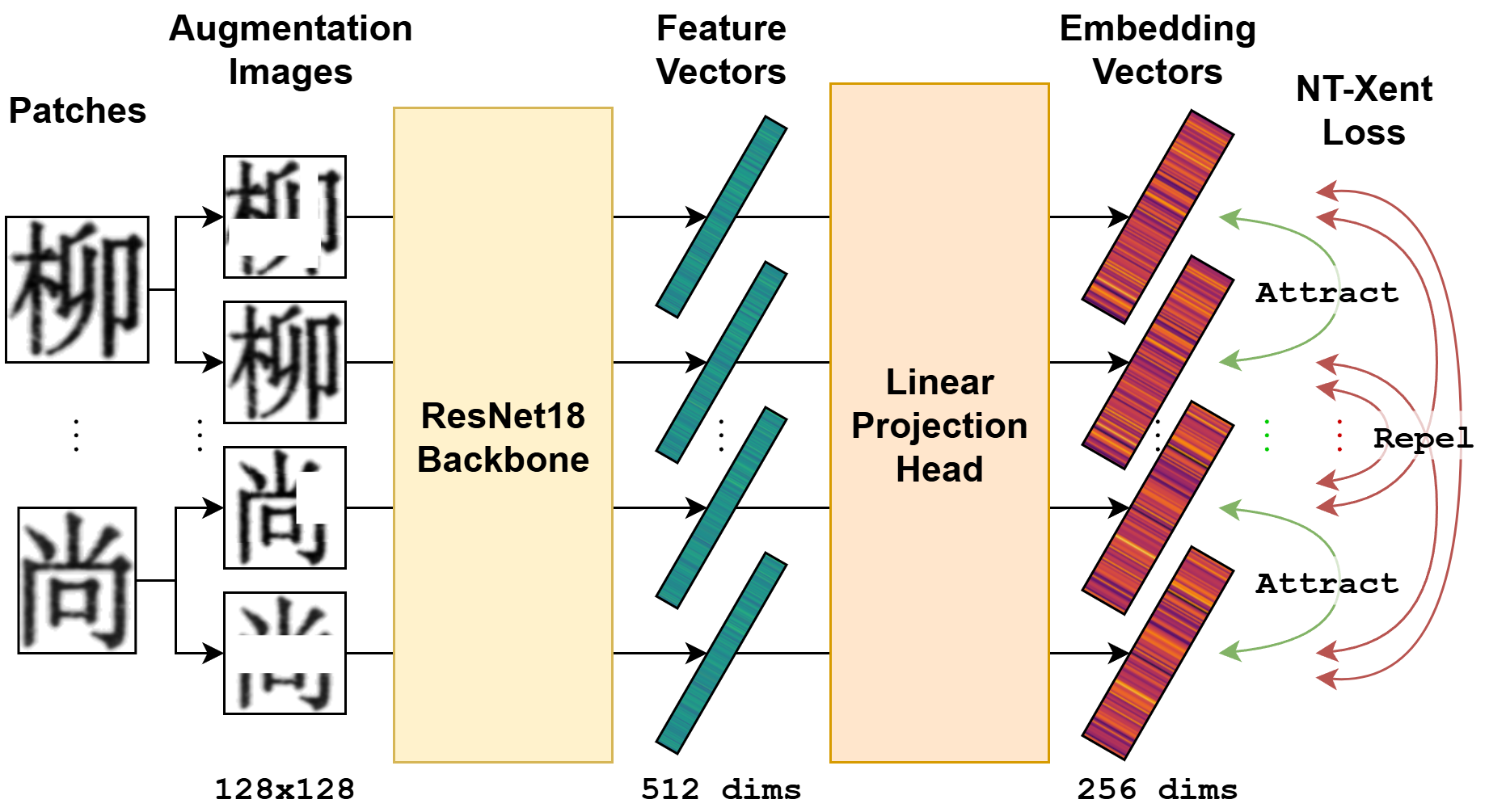}
        \vspace{-8pt}
        \caption{The SimCLR-based self-supervised learning framework for feature extraction.}
        \label{fig:fig07_simclr}
    \end{figure}

    We use a SimCLR \cite{simclr}-based self-supervised learning framework to extract robust feature representations of character patches without requiring manual annotations. The ResNet18 \cite{resnet} backbone encoder, adapted from ImageNet \cite{imagenet}-pretrained weights, along with a linear projection head, maps patches into a hyper-spherical embedding space. For each publication, all extracted candidate patches (e.g., 205,712 patches in Volume \textit{Jiangsu II}, \textit{The Anthology}) are used for training. Positive pairs are generated by applying random data augmentations—such as scaling, rotation, color jittering, and random erasing. NT-Xent loss \cite{simclr} is employed to attract positive pairs and repel negative ones. Fig.~\ref{fig:fig07_simclr} illustrates the training session diagram\footnote{This figure is redrawn based on an illustration from a blog article of Google Research: \\\url{https://research.google/blog/advancing-self-supervised-and-semi-supervised-learning-with-simclr/}}.

    \vspace{-2pt}
    
    The embedding similarity metric between two patches is computed as their cosine similarity in the embedding space.

    \subsubsection{Summary of Character Matching}
    
\begin{figure}[h!]
        \centering
        \includegraphics[width=1\columnwidth]{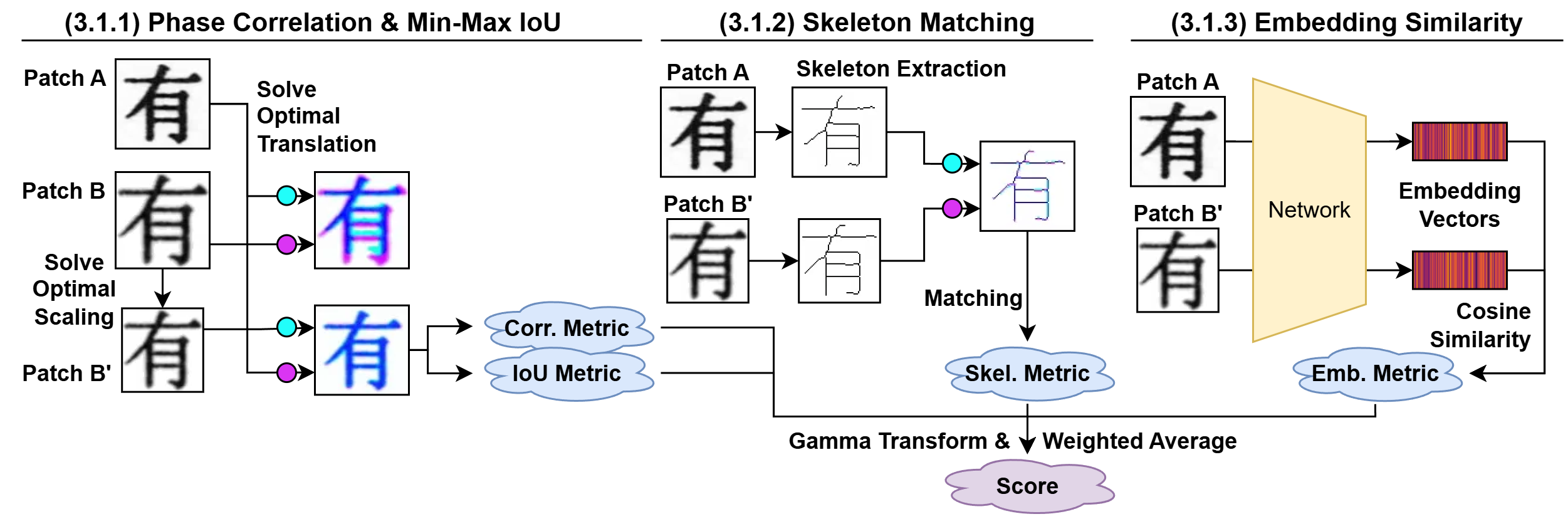}
        \vspace{-8pt}
        \caption{Summary of the four similarity metrics used in Chinese lyric patch matching.}
        \label{fig:fig08_similarity}
    \end{figure}

    The overall similarity score is calculated as a weighted average of the four metrics above, with gamma adjustments. A summary of these four metrics is shown in Fig.~\ref{fig:fig08_similarity}.

    \vspace{-6pt}

    \subsection{Template-based Lyric OCR}

    Lyric patch candidates are extracted using adaptive connected component analysis with multi-threshold binarization, capturing character strokes under varying conditions. The components are then merged and filtered using rules designed for \textit{Hanzi} characters, with R-tree indexing for faster merging.

\begin{figure}[h!]
        \centering
        \includegraphics[width=0.64\columnwidth]{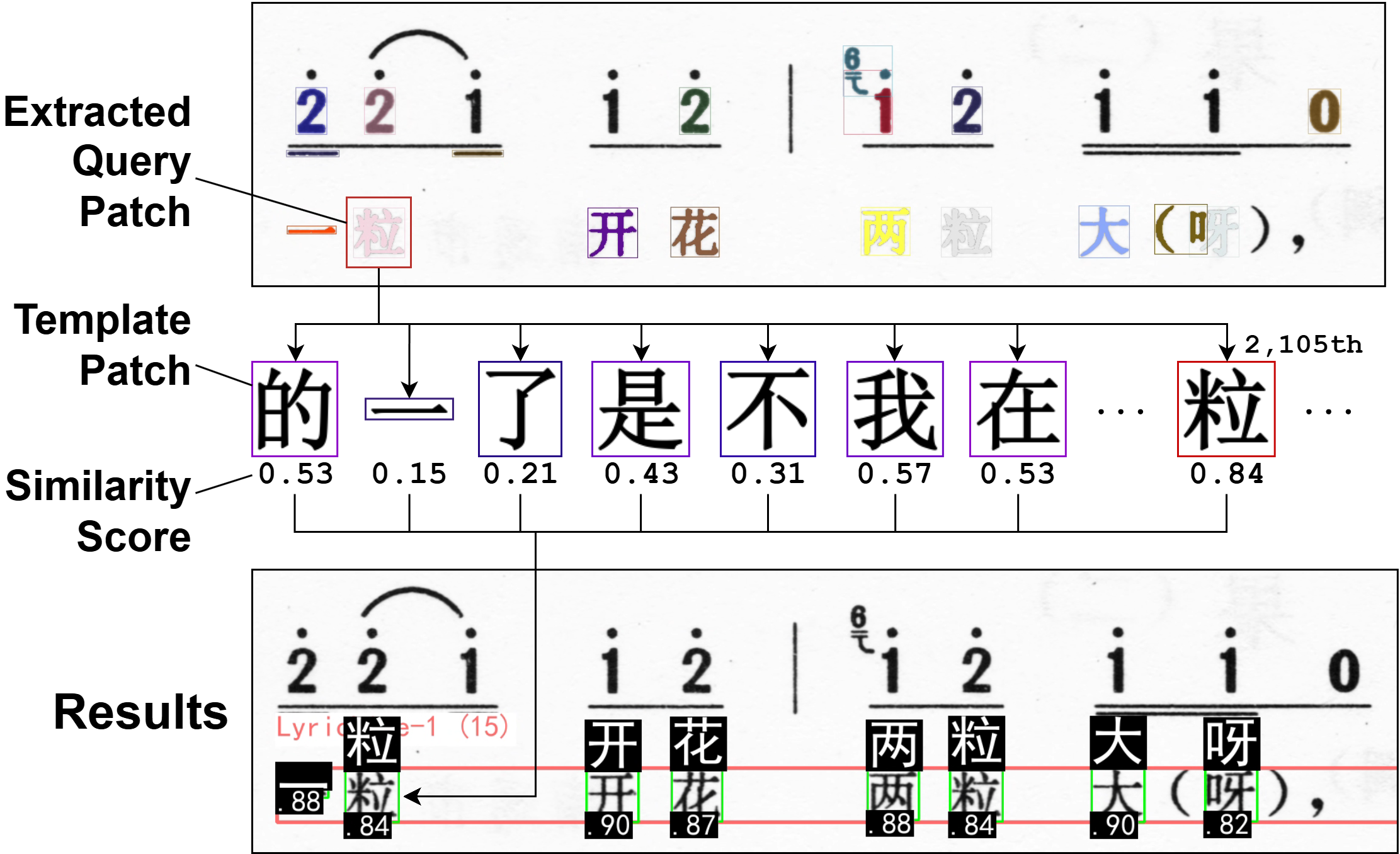}
        \vspace{-8pt}
        \caption{Example of template-based lyric OCR.}
        \label{fig:fig09_ocr_template_matching}
    \end{figure}

    A \textit{Hanzi} template table of 14,975 patches was created from a frequency-ordered character list\footnote{Hongbing Xing, Chinese character list from 2.5 billion words corpus ordered by frequency:\\\url{https://faculty.blcu.edu.cn/xinghb/zh_CN/article/167473/content/1437.htm}}, with font and size matching the target publication (e.g., SimSun and SimHei for volumes in \textit{The Anthology}). For each extracted patch, similarity scores with \textit{Hanzi} template patches are computed, and the best match predicts the character (see Fig.~\ref{fig:fig09_ocr_template_matching}). Half-resolution matching and frequency data are used for pruning irrelevant templates.

    \vspace{-12pt}

    \subsection{Validation on Lyric OCR}

\begin{figure}[htb]
        \centering
        \includegraphics[width=1\columnwidth]{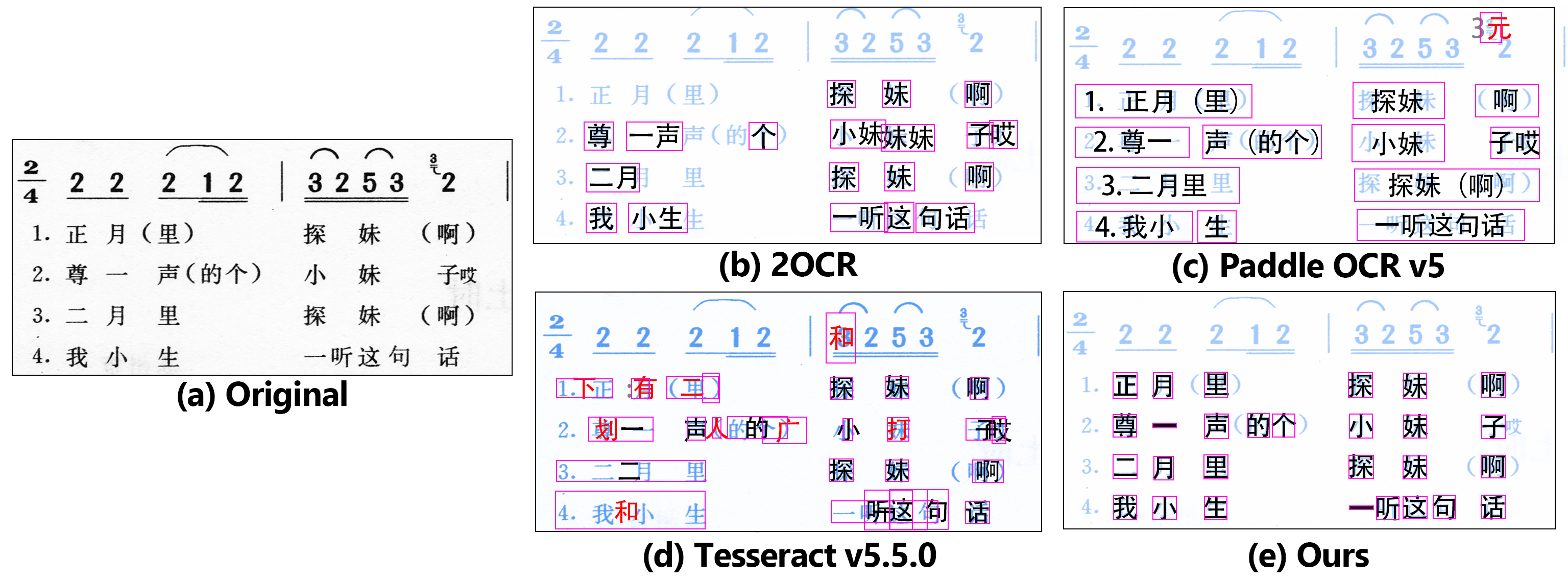}
        \vspace{-16pt}
        \caption{Comparison of OCR results from different toolkits. In (b)$\sim$(e), the original image is overlaid in blue for comparison.}
        \label{fig:fig10_ocr_validation}
    \end{figure}

    We evaluated several modern OCR toolkits, including 2OCR\footnote{2OCR: \url{https://2ocr.com/}}, Paddle OCR \cite{paddle_ocr}, Tesseract\footnote{Tesseract: \url{https://github.com/tesseract-ocr/tesseract}}, and our proposed method, on randomly selected pages from Volume \textit{Jiangsu II}, \textit{The Anthology}, which contains over 500 characters (as shown in Fig.\ref{fig:fig10_ocr_validation}). The results are summarized in Tab.\ref{tab:ocr_validation}. Incorrectly identified characters are counted twice: once as false positives (FP) and once as false negatives (FN).

    \begin{table}[htb]
        \small
        \centering
        \caption{Comparison of OCR results from different toolkits.}
        \vspace{-6pt}
        \begin{tabular}{c|cccc|c}
            \toprule[1pt]
            \textbf{Metric} & \textbf{TP} & \textbf{FN} & \textbf{FP} & \textbf{F1-Score} & \textbf{Bounding Box Type} \\
            \midrule[0.5pt]
            2OCR & 535 & 15 & 6 & 0.981 & Region-wise \\
            \midrule[0.5pt]
            Paddle OCR v5 & 543 & 6 & 3 & \textbf{0.992} & Region-wise \\
            \midrule[0.5pt]
            Tesseract v5.5.0 & 296 & 255 & 214 & 0.558 & Region-wise \\
            \midrule[0.5pt]
            Ours & 533 & 15 & 8 & 0.979 & \textbf{Character-wise} \\
            \bottomrule[1pt]
        \end{tabular}\\
        \label{tab:ocr_validation}
    \end{table}

    \vspace{-10pt}

    The results indicate that our method achieves performance comparable to modern OCR tools while providing more precise localization of individual characters. We observe that our method occasionally struggles with distinguishing between visually similar characters, such as ``待"/``侍", ``糯"/``懦", ``唷"/``啃" and ``家"/``冢". We plan to collaborate with natural language models, which can leverage contextual information and correct these errors.

    \vspace{-4pt}

    \section{Dataset Construction and Validation on \textit{The Anthology of Chinese Folk Songs}}

    \textit{The Anthology of Chinese Folk Songs} is a comprehensive collection of over 30,000 folk songs gathered from various regions across China. Compiled into 31 volumes, this anthology was published in the 1980s and 1990s and transcribed using \textit{Jianpu}. This archive plays a vital role in the study of regional cultural research. In this section, we describe the process of constructing a digital dataset based on \textit{The Anthology}, which includes both melody and lyric data.

    We employed the algorithms outlined in the previous sections\footnote{Our code is available at \url{https://github.com/m-july/Renaissance-of-Expert-Systems-Jianpu-OMR-Pipeline-v251103}.} to extract and convert the musical data from the printed scores into MusicXML and MIDI formats. The resulting dataset\footnote{Our dataset is available at \url{https://github.com/m-july/Anthology-of-Chinese-Folk-Songs-v251103}.} is provided in two versions:

    \vspace{-4pt}
    \begin{enumerate}

    \item \textbf{Melody-only dataset} (file per volume): Contains only the melody (excluding lyrics and metadata), comprising the merged melody from each of the selected 8 volumes, with a total of over 5,000 songs and over 300,000 notes.

    \vspace{-4pt}
    
    \item \textbf{Curated subset} (file per song): Includes melody, lyrics and  metadata, derived from \textit{Jiangsu I and II} volumes, containing over 1,400 songs and over 100,000 notes.

    \end{enumerate}
    
    \vspace{-8pt}

    \subsection{Dataset Construction Efficiency}
    
    Our testing environment consists of a 16-core Intel i7-13700KF CPU @ 3.40GHz, an NVIDIA GeForce RTX 4090 GPU (24GB VRAM), and 64GB of RAM. Efficiency tests were performed on Volume \textit{Jiangsu II}, which consists of 646 pages and over 600 songs.
    
    The melody-only extraction process took 3.50 hours (40.3\% detecting digits, 45.0\% detecting other musical elements, 14.7\% relational analysis and format conversion). On average, each page took 19.5 seconds to process, resulting in the detection of 268,531 musical elements, including 53,483 notes.
    
    The lyric recognition algorithm prioritizes precision and alignment with the notes over throughput. As a result,  the average recognition time per character was 6.5 seconds, translating to 10.5 minutes per page on average.

    \vspace{-4pt}

    \subsection{Dataset Comparison}

    To assess the scale of our dataset, we compared it to other prominent Chinese music datasets in terms of size and format availability. As shown in Tab.~\ref{tab:comparison_of_datasets}, our dataset stands out for its diverse collection of folk songs, representing one of the largest datasets available for musicological analysis, especially considering the inclusion of lyrics.

    \vspace{-2pt}

    \begin{table}[htb]
        \small
        \centering
        \caption{Comparison of prominent Chinese music datasets.}
        \vspace{-6pt}
        \begin{tabular}{c|cc|cc|c}
            \toprule[1pt]
            \textbf{Dataset} & \textbf{Genre} & \textbf{Notation} & \textbf{Entries} & \textbf{Notes} & \textbf{Lyric}\\
            \midrule[0.5pt]
            POP909 \cite{pop909}& Pop & MIDI & 2,898 & 4,812,678 & None \\
            \midrule[0.5pt]
            \makecell{CCMusic \cite{ccmusic}\\Pop Database$^{*}$} & Pop & \makecell{MusicXML \&\\MIDI \& WAV} & >100 & >10,000 & Aligned \\
            \midrule[0.5pt]
            \textit{Jiugong Dacheng} \cite{jiugong_dacheng} & Traditional & MusicXML \& MIDI & 6,556 & 701,586 & Aligned \\
            \midrule[0.5pt]
            Essen's Chinese Subset & Folk & Kern (ABC-like) & 2,179 & 109,613 & None \\
            \midrule[0.5pt]
            Ours (Melody-only) & Folk & MusicXML \& MIDI & >5,000 & >300,000 & None \\
            \midrule[0.5pt]
            Ours (Curated) & Folk & MusicXML \& MIDI & >1,400 & >100,000 & Aligned \\
            \bottomrule[1pt]
        \end{tabular}\\

        \vspace{0.5em}

        \footnotesize{$^{*}$Midi-Wav Bi-directional Database, CCMusic: \url{https://ccmusic-database.github.io/database/cpop.html}}\\
        
        \label{tab:comparison_of_datasets}
    \end{table}

    \vspace{-14pt}

    \subsection{\textit{Jianpu} OMR Pipeline Validation}

    We evaluated the recognition performance of the pipeline on random-selected pages (including 233 notes). Several evaluation metrics were calculated on: (a) Note/rest-wise: digit detection, pitches and durations; (b) Measure-wise: measure length (the sum of note durations); (c) Lyrics: lyric detection and recognition.

    \vspace{-6pt}
    
    \begin{figure}[h]
        \centering
        \includegraphics[width=0.9\columnwidth]{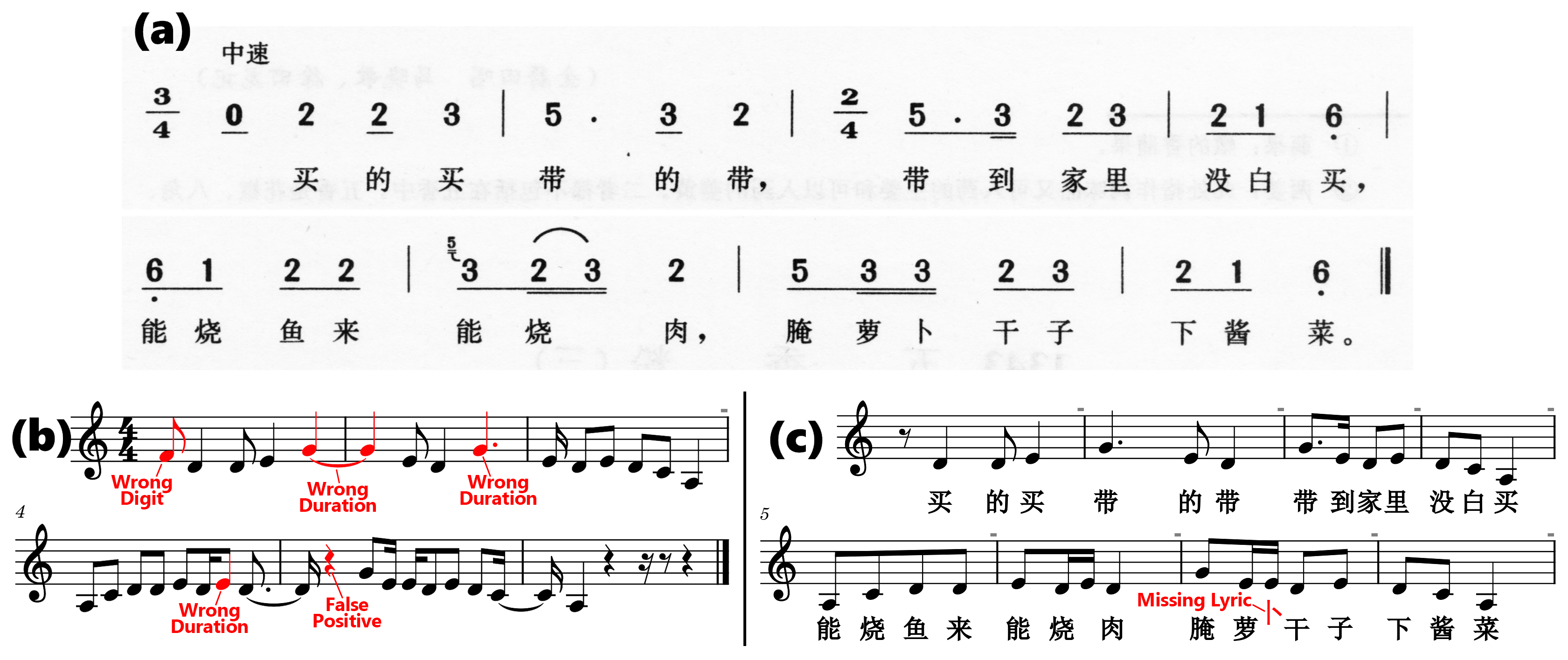}
        \vspace{-12pt}
        \caption{Comparison of OMR results from different pipelines. (a) Original score image; (b) Result of \cite{bu_fan_jianpu_omr}; (c) Result of ours.}
        \label{fig:fig11_omr_validation}
    \end{figure}

    \begin{table}[h]
        \small
        \centering
        \caption{Comparison of existing \textit{Jianpu} OMR pipelines.}
        \makebox[\textwidth][c]{
            \begin{tabular}{c|cccc|c|cc}
                \toprule[1pt]
                \textbf{Type} & \multicolumn{4}{c|}{\textbf{Note/rest}} & \textbf{Measure} & \multicolumn{2}{c}{\textbf{Lyric}}\\
                \midrule[0.5pt]
                \makecell{\textbf{Metric}\\F1/Acc.\\\footnotesize{(Details)}} & \makecell{\textbf{Detection}\\F1-score\\\footnotesize{(TP/FN/FP)}} & \makecell{\textbf{Pitch}\\Accuracy\\\footnotesize{(T/F)}} & \makecell{\textbf{Duration}\\Accuracy\\\footnotesize{(T/F)}} & \makecell{\textbf{Joint$^a$}\\F1-score\\\footnotesize{(TP/FN/FP)}} & \makecell{\textbf{Length}\\Accuracy\\\footnotesize{(T/F)}} & \makecell{\textbf{Detection}\\F1-score\\\footnotesize{(TP/FN/FP)}} & \makecell{\textbf{Joint$^a$}\\F1-score\\\footnotesize{(TP/FN/FP)}} \\
                \midrule[0.5pt]
                \makecell{In \cite{wang_qi_jianpu_omr}$^b$\\\footnotesize{(CNNs)}} & \makecell{1.000$^b$\\\footnotesize{(Assumed)}} & \makecell{94.8\%$^b$\\\footnotesize{(221/12)}} & \makecell{91.0\%$^b$\\\footnotesize{(212/21)}} & \makecell{0.858$^b$\\\footnotesize{(200/33/33)}} & \makecell{81.1\%$^b$\\\footnotesize{(43/10)}} & N/A & N/A \\
                \midrule[0.5pt]
                \makecell{In \cite{bu_fan_jianpu_omr}\\\footnotesize{(YOLO-like)}} & \makecell{0.969\\\footnotesize{(233/0/15)}} & \makecell{91.0\%\\\footnotesize{(212/21)}} & \makecell{88.8\%\\\footnotesize{(207/26)}} & \makecell{0.785\\\footnotesize{(188/44/59)}} & \makecell{62.3\%\\\footnotesize{(33/20)}} & N/A & N/A \\
                \midrule[0.5pt]
                \makecell{Ours\\\footnotesize{(Expert-system)}} & \makecell{\textbf{1.000}\\\footnotesize{(233/0/0)}} & \makecell{\textbf{100.0\%}\\\footnotesize{(233/0)}} & \makecell{\textbf{95.3\%}\\\footnotesize{(222/11)}} & \makecell{\textbf{0.951}\\\footnotesize{(222/12/11)}} & \makecell{\textbf{84.9\%}\\\footnotesize{(45/8)}} & \makecell{\textbf{0.954}\\\footnotesize{(324/31/0)}} & \makecell{\textbf{0.931}\\\footnotesize{(316/39/8)}} \\
                \bottomrule[1pt]
            \end{tabular}
        }\\
        \vspace{0.5em}

        \footnotesize{$^{a}$ \textbf{Joint} F1: evaluates detection and content recognition simultaneously. A mis-recognized symbol is counted twice—once as a false negative (missed truth) and once as a false positive (spurious prediction).}\\
        
        \footnotesize{$^{b}$ Scores for \cite{wang_qi_jianpu_omr} are from a theoretical simulation on the test set, using the per-class recognition accuracies reported by the paper under an independence assumption, with perfect detection and localization assumed.}\\

        \label{tab:comparison_of_omrs}
    \end{table}

    These metrics were compared to results from established \textit{Jianpu} OMR pipelines, as reported in two recent papers. The results are summarized in Tab.~\ref{tab:comparison_of_omrs} and an example is illustrated in Fig.~\ref{fig:fig11_omr_validation}. Our method demonstrated strong performance, achieving a note-wise F1-score of 0.951. In terms of length consistency across musical measures, our approach achieved an accuracy of 84.9\%. Notably, our method is the only one that simultaneously performs lyric recognition, achieving a high recognition F1-score of 0.931.

    \section{Conclusion}

    This study presented a modular, expert-system pipeline that converts printed Chinese \textit{Jianpu} scores with lyrics into structured MusicXML/MIDI, digitizing more than 5,000 songs from \textit{The Anthology of Chinese Folk Songs}. The results show that the proposed approach attains note, rhythm, and lyric recognition with high accuracy while using only minimal annotated data.

    Our system demonstrates how traditional computer-vision techniques can be powerfully combined with modern deep-learning components. The classical algorithms supply strong geometric priors while the neural modules inject robustness and unsupervised feature compression; this yields a data-efficient, high-precision recognizer that would be hard to achieve with either paradigm alone.

    The resulting resources enable two immediate avenues of future work. The constructed melody-and-lyric corpus can serve as training material for downstream tasks and models, such as MusicBERT \cite{music_bert}. Moreover, with minor changes, the recognition modules themselves can act as a ready-made benchmark for OMR researchers, effectively turning printed \textit{Jianpu} scores into high-quality data for broader music-AI exploration.

    \section*{Appendix}

    \subsection*{A.1  Key modification to KD-tree pruning}

    Each node stores an axis-aligned bounding box (AABB). For a query point $\boldsymbol p$, we compute an admissible lower bound $\underline d$ on the anisotropic distance to the box $\mathcal B$:

    \vspace{-8pt}

    \begin{equation}
        \underline d(\boldsymbol p,\mathcal B) = \sqrt{(s_x\,\delta_x)^2 + (s_y\,\delta_y)^2}
    \end{equation}
    
    where $\delta_x,\delta_y$ are clamped distances from $\boldsymbol p$ to the box. This scaling ensures correct pruning under elliptical distance. The search proceeds in a best-first manner using a priority queue keyed by $\underline d$, guaranteeing optimal nearest-neighbor results.

    \subsection*{A.2  Normalization of Image Patch Similarity}

    \textbf{Phase Correlation.} Denoting energies $E_1 = \sum_{x, y} f_1^2(x, y)$ and $E_2 = \sum_{x, y} f_2^2(x, y)$. Applying the Cauchy–Schwarz inequality, it follows that $|C(x, y)| \le \sqrt {E_1 E_2}$. Thus $C$ can be normalized into $[-1, 1]$ by the factor $1/\sqrt{E_1 E_2}$.
    
    \textbf{Min-max IoU} has a natural range of $[0, 1]$.
    
    \textbf{Skeleton Matching.} The worst-case scenario happens when all pairs are unmatched, yielding an upper bound $J_{\uparrow} = \lambda^2 \cdot (\frac {m+n}{2})$. Thus $J$ can be normalized into $[0, 1]$ by the factor $1/J_\uparrow$ regardless of $\lambda$, $m$ and $n$. Transform $y=\exp (-x)$ is adopted to gain a decaying similarity score ranged $[1/{\rm e}, 1]$.
    
    \textbf{Embedding Similarity.} Cosine similarity has a natural range of $[-1, 1]$.

    \bibliographystyle{gbt-7714-2015-numerical}
    \bibliography{mybib}
	
\end{document}